\documentclass[letterpaper]{article}

\usepackage{aaai}
\usepackage{times}
\usepackage{helvet}
\usepackage{courier}
\usepackage{graphicx}
\usepackage{subcaption}
\usepackage{amsfonts}
\usepackage{amsmath}
\usepackage{textcomp}

\newcommand{\R}{\mathbb{R}}

\begin{document}

\title{\textit{Sequence-to-Set} Semantic Tagging: End-to-End Multilabel Prediction using Neural Attention for Complex Query Reformulation and Automated Text Categorization}
 %\author{author\\address}
 \author{
    Manirupa Das\textsuperscript{$\dagger$}, 
    Juanxi Li\textsuperscript{$\dagger$},  
    Eric Fosler-Lussier\textsuperscript{$\dagger$},  
    Simon Lin\textsuperscript{$\ddagger$},  \\
    {\Large \bf Soheil Moosavinasab\textsuperscript{$\ddagger$}},
    {\Large \bf Steve Rust\textsuperscript{$\ddagger$}},  
    {\Large \bf Yungui Huang\textsuperscript{$\ddagger$}} \&
    {\Large \bf Rajiv Ramnath\textsuperscript{$\dagger$}}\\
    {\textsuperscript{$\dagger$}\Large The Ohio State University} \& 
    {\textsuperscript{$\ddagger$}\Large Nationwide Children's Hospital} \\
    {\tt \{das.65, li.8767, fosler-lussier.1, ramnath.6\}@osu.edu } \\
    {\tt \{Simon.Lin, Steve.Rust, Yungui.Huang\}@nationwidechildrens.org } \\
  }
  
%\author{AuthorOne},\textsuperscript{1}
%\author{AuthorTwo},\textsuperscript{2}
%\author{AuthorThree},\textsuperscript{3}
%\author{AuthorFour},\textsuperscript{4}
%\author{AuthorFive}, \textsuperscript{5}\\
%\textsuperscript{1}AffiliationOne}\\
%\textsuperscript{2}AffiliationTwo}\\
%\textsuperscript{3}AffiliationThree}\\
%\textsuperscript{4}AffiliationFour}\\
%\textsuperscript{5}AffiliationFive}\\
\maketitle

\begin{abstract}
\begin{quote}
Novel contexts may often arise in complex querying scenarios such as in evidence-based medicine (EBM) involving biomedical literature, that may not explicitly refer to entities or canonical concept forms occurring in any fact- or rule-based knowledge source such as an ontology like the UMLS. Moreover, hidden associations between candidate concepts meaningful in the current context, may not exist within a single document, but within the collection, via alternate lexical forms. %Our objective therefore, is to implicitly learn and encode the novel contexts that may relate two or more documents by learning the best possible intrinsic document representations, incorporating both local contextual information on word usage occurring within a document, as well as global information on term usage occurring across the document collection. 
%We aim to achieve this by using word embeddings trained on the whole corpus in our models, within a novel \textit{sequence-to-set} pipeline,  that offers a generic paradigm for unsupervised semantic tagging of documents in any domain. 
Thus, to predict such semantically related concept tags, and inspired by the recent success of sequence-to-sequence neural models in delivering the state-of-the-art in a wide range of NLP tasks, %and building upon the successes of previous neural generalized language model--based methods, 
we develop a novel \textit{sequence-to-set} framework with neural attention for learning document representations that can effect term transfer within the corpus, for semantically tagging a large collection of documents. 

We demonstrate that our proposed method can be effective in
both a supervised multi-label classification setup for text
categorization, as well as in a unique unsupervised setting with no human-annotated document labels that uses no external knowledge resources and only corpus-derived term statistics to drive the training. Further, we show that semi-supervised training using our architecture on large amounts of unlabeled data can augment performance on the text categorization task when limited labeled data is available. % We evaluate our method on supervised, semi-supervised and unsupervised setups of semantic tagging of documents via multi-label classification from document collections having various levels of human-derived tag labels. 
Our approach to generate document encodings employing our sequence-to-set models for inference of semantic tags, gives to the best of our knowledge, the state-of-the-art for both, the \textbf{unsupervised} query expansion task for the \textbf{TREC CDS 2016} challenge dataset when evaluated on an Okapi BM25--based document retrieval system; and also over the MLTM baseline \cite{soleimani2016semi}, for both \textbf{supervised} and \textbf{semi-supervised} multi-label prediction tasks on the \textbf{del.icio.us} and \textbf{Ohsumed} datasets. We will make our code and data publicly available.

%We evaluate our framework on the \textbf{TREC 2016 clinical decision support challenge} dataset having no human-derived document labels for the unsupervised task; on a collection of tagged documents from the \textbf{del.icio.us} folksonomy for the supervised task; as well as a tagged collection of medline abstracts from the \textbf{Ohsumed} dataset for the semi-supervised task, using only a fraction of the labeled data for training and majority of the data without human-annotated tags. %, evaluating the inferred semantic tags from this setup via complex query reformulation of clinical queries in this same dataset, in a pseudo-relevance feedback-based query expansion setting. 
%Our approach to generate document encodings employing our sequence-to-set models for inference of semantic tags, gives to the best of our knowledge, the state-of-the-art when evaluated on the TREC challenge dataset query expansion, on top of a standard Okapi BM25--based document retrieval system for the unsupervised task, and statistically significant AUC over the baseline, for both the supervised and semi-supervised tasks for the del.icio.us and Ohsumed datasets.
\end{quote}
\end{abstract}

\section{Introduction}

Recent times have seen an upsurge in efforts towards  personalized medicine where clinicians tailor their medical decisions to the individual patient, based on the patient's genetic information, other molecular analysis, and the patient's preference. This often requires them to combine clinical experience with evidence from scientific research, such as that available from biomedical literature, in a process known as evidence-based medicine (EBM). Finding the most relevant recent research however, is challenging not only due to the volume and the pace at which new research is being published, but also due to the complex nature of the information need,  arising for example, out of a clinical note which may be used as a query. 
This calls for better automated methods for natural language understanding (NLU), e.g. to derive a set of key terms or \textit{related concepts} helpful in appropriately transforming a complex query, by reformulation so as to be able to handle and possibly resolve medical jargon, lesser-used acronyms, misspelling, multiple subject areas and often multiple references to the same entity or concept, and retrieve the most related, yet most comprehensive set of useful results.

%Finding the most relevant  recent research however, is not only challenging due to the volume and the pace at which new research is being published, but also due to the complex nature of the information need,  arising for example, out of a clinical note which may be used as a query. This calls for better automated methods for natural language understanding (NLU), e.g. to derive a set of key terms or \textit{related concepts} helpful in appropriately transforming a complex query, by reformulation. Here the goal is to be able to handle and possibly resolve medical jargon, lesser-used acronyms, misspelling, multiple subject areas and often multiple references to the same entity or concept, and retrieve the most related, yet most comprehensive set of useful results.

At the same time, tremendous strides have been made by recent neural machine learning models in reasoning with texts on a wide variety of NLP tasks. In particular, sequence-to-sequence (seq2seq) neural models often employing \textit{attention} mechanisms,  have been largely successful in delivering the state-of-the-art for tasks such as  machine translation \cite{bahdanau2014neural}, \cite{vaswani2017attention}, handwriting synthesis \cite{graves2013generating}, image captioning \cite{xu2015show}, speech recognition \cite{chorowski2015attention} and document summarization \cite{cheng2016neural}. Inspired by these successes, we aimed to harness the power of sequential \textit{encoder-decoder} architectures with attention, to train end-to-end differentiable models that are able to learn the best possible representation of input documents in a collection while being predictive of a set of \textit{key terms} that best describe the document. These will be later used to \textit{transfer} a relevant but diverse set of key terms from the most related documents, for ``semantic tagging'' of the original input documents so as to aid in downstream query refinement for IR by pseudo-relevance feedback \cite{xu2000improving}.

To this end and to the best of our knowledge, we are the first to employ a novel, completely unsupervised end-to-end neural attention-based document representation learning approach, using no external labels, in order to achieve the most meaningful term transfer between related documents, i.e. semantic tagging of documents, in a ``pseudo-relevance feedback''--based \cite{xu2000improving} setting for unsupervised query expansion.   This may also be seen as a method of document expansion as a means for obtaining query refinement terms for downstream IR. The following sections give an account of our specific architectural considerations in achieving an end-to-end neural framework for semantic tagging of documents using their representations, and a discussion of the results obtained from this approach.

 %detailed in following sections.
%We aim to achieve this using word embeddings trained on the whole corpus in our models, within a novel sequence-to-set pipeline. Thus, \textit{Seq2set} offers a generic paradigm for unsupervised semantic tagging of documents in any domain towards automated text categorization for information filtering and generating search recommendations.

\section{Related Work}

Pseudo-relevance feedback (PRF), a \textit{local context analysis} method for automatic query expansion (QE), is extensively studied in information retrieval (IR) research as a means of addressing the word mismatch between queries and documents. It adjusts a query relative to the documents that initially appear to match it, with the main assumption that the top-ranked documents in the first retrieval result contain many useful terms that can help discriminate relevant documents from irrelevant ones \cite{xu2000improving}, \cite{cao2008selecting}. It is motivated by \textit{relevance feedback} (RF), a well-known IR technique that modifies a query based on the relevance judgments of the retrieved documents \cite{salton1990application}. It typically adds common terms from the relevant documents to a query and re-weights the expanded query based on term frequencies in the relevant documents relative to the non-relevant ones. Thus in PRF we find an initial set of most relevant documents, then assuming that the top \textit{k} ranked documents are relevant, RF is done as before, without manual interaction by the user.  The added terms are, therefore, common terms from the top-ranked documents.

To this end, \cite{cao2008selecting} employ term classification for retrieval effectiveness, in a ``supervised'' setting, to select most relevant terms. \cite{palangi2016deep} employ a deep sentence embedding approach using LSTMs and show improvement over standard sentence embedding methods, but as a means for directly deriving encodings of queries and documents for use in IR, and not as a method for QE by PRF. In another approach, \cite{xu2017learning} train autoencoder representations of queries and documents to enrich the feature space for learning-to-rank, and show gains in retrieval performance over pre-trained rankers. But this is a fully supervised setup where the queries are \textit{seen} at train time. \cite{pfeiffer2018neural} also use an autoencoder-based approach for actual query refinement in pharmacogenomic document retrieval. However here too, their document ranking model uses the encoding of the query and the document for training the ranker, hence the queries are \textit{not unseen} with respect to the document during training. They mention that their work can be improved upon by the use of seq2seq-based approaches. In this sense, i.e. with respect to QE by PRF and learning a sequential document representation for document ranking, our work is most similar to \cite{pfeiffer2018neural}. However the queries are completely unseen in our case and we use only the documents in the corpus, to train our neural document language models from scratch in a completely unsupervised way. 

Classic sequence-to-sequence models like \cite{sutskever2014sequence} demonstrate the strength of recurrent models such as the LSTM in capturing short and long range dependencies in learning effective encodings for the end task. Works such as \cite{graves2013generating}, \cite{bahdanau2014neural}, \cite{rocktaschel2015reasoning}, further stress the key role that attention, and multi-headed attention \cite{vaswani2017attention} can play in solving the end task. We use these insights in our work.

According to the detailed report provided for this dataset and task in \cite{overviewcds2016} all of the systems described perform \textbf{direct query reweighting} aside from \textbf{supervised term expansion} and are highly tuned to the clinical queris in this dataset. In a related medical IR challenge \cite{roberts2017information} the authors specifically mention that with only six partially annotated queries for system development, it is likely that systems were either under- or over-tuned on these queries. Since the setup of the seq2set framework is an attempt to model the PRF based query expansion method of its closest related work \cite{das2018phrase2vecglm} where the effort is also to train a neural generalized language model for unsupervised semantic tagging, we choose this system as the benchmark to compare against to our end-to-end approach for the same task.

\section{Methodology}

\begin{figure*}[t]
    \centering
    \includegraphics[width=0.94\textwidth]{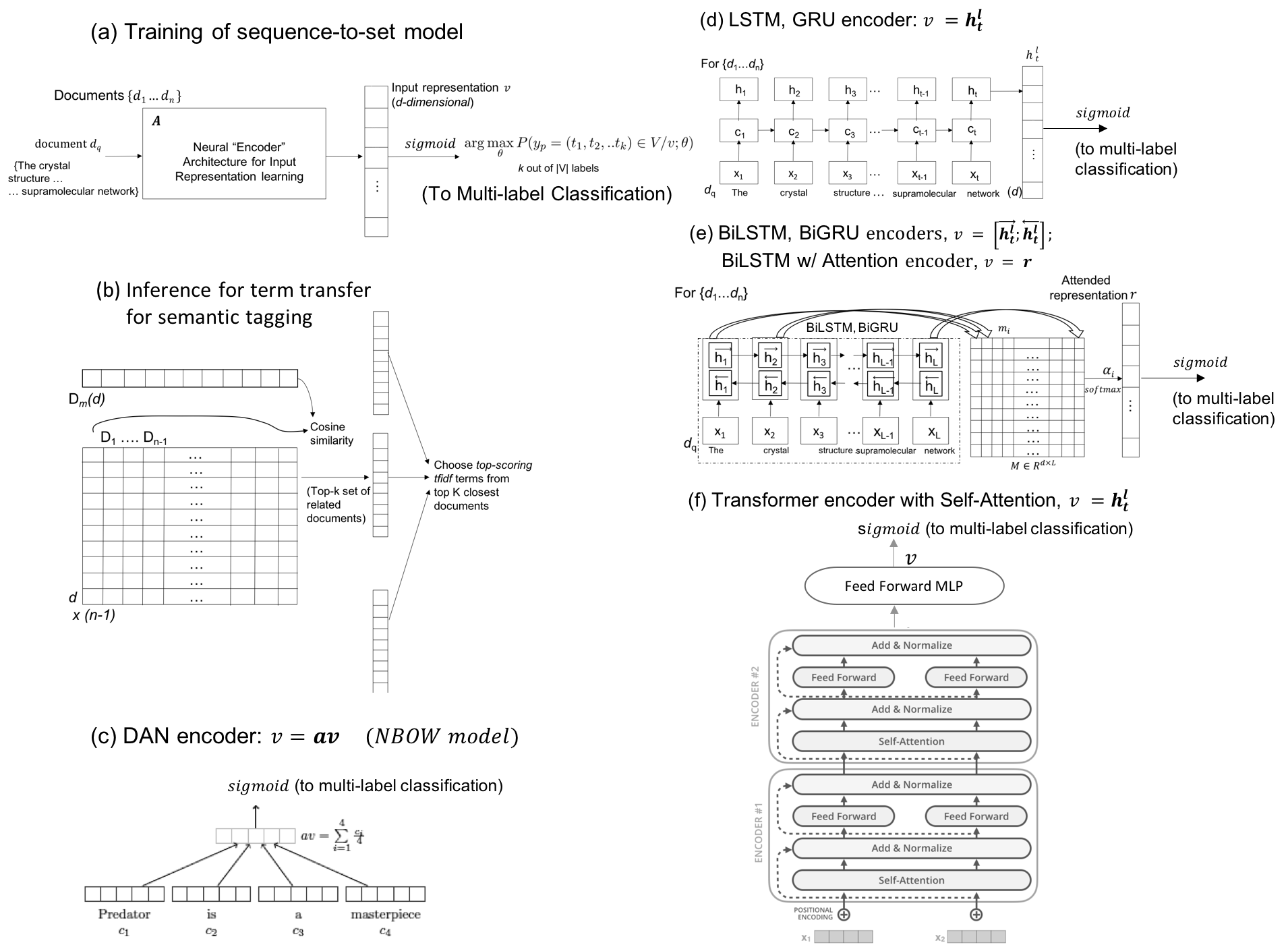}
    \caption{Overview of \textit{Sequence-to-Set} Framework. (a) Method for training document or query representations, (b) Method for Inference via \textbf{term transfer} for semantic tagging; Document Sequence Encoders: (c) Deep Averaging encoder; (d) LSTM last hidden state, GRU encoders; (e) BiLSTM last hidden state, BiGRU (shown in dotted box),  BiLSTM attended hidden states encoders; and (f) Transformer self-attentional encoder [source: \cite{alammartransformer2018}].}
    \label{seq2set_fig}
\end{figure*}

Drawing on sequence-to-sequence modeling approaches for text classification, e.g. textual entailment \cite{rocktaschel2015reasoning} and machine translation \cite{sutskever2014sequence}, \cite{bahdanau2014neural}  we adapt from these settings into a \textit{sequence-to-set} framework, for learning  representations of input documents, in order to derive a meaningful set of terms, or  \textit{semantic tags} drawn from a closely related set of documents, that expand the original documents. These document expansion terms are then used downstream for query reformulation via PRF, for unseen queries. We employ an end-to-end framework for unsupervised representation learning of documents using TFIDF-based \textit{pseudo-labels} (Figure \ref{seq2set_fig}(a))and a separate cosine similarity-based ranking module for semantic tag inference (Figure \ref{seq2set_fig}(b)). 

We employ various methods such as doc2vec, Deep Averaging, sequential models such as LSTM, GRU, BiGRU, BiLSTM, BiLSTM with Attention and Self-attention, detailed in Figure \ref{seq2set_fig}(c)-(f), for learning fixed-length input document representations in our framework. We apply methods like DAN \cite{iyyer2015deep}, LSTM, and BiLSTM as our baselines and formulate attentional models including a self-attentional Transformer-based one \cite{vaswani2017attention} as our proposed augmented document encoders.

Further, we hypothesize that a sequential, bi-directional or attentional encoder coupled with a decoder, i.e. a sigmoid or softmax prediction layer, that conditions on the encoder output $v$ (similar to an approach by \cite{kiros2015skip} for learning a neural probabilistic language model), would enable learning of the optimal semantic tags in our unsupervised query expansion setting, while  modeling directly for this task in an end-to-end neural framework. In our setup the decoder predicts a meaningful set of concept tags that best describe a document according to the training objective. The following sections describe our setup.
%\newpage
\subsection{The \textit{Sequence-to-Set} Semantic Tagging Framework}
\textit{\textbf{Task Definition}}: For each query document $d_q$ in a given a collection of documents $D = \{d_1, d_2,..., d_N\}$, represented by a set of $k$ keywords or labels, e.g. $k$ terms in $d_q$ derived from $top-|V|$ TFIDF-scored terms, find an \textit{\textbf{alternate}} set of $k$ most relevant terms coming from documents \textit{“most related”} to $d_q$ from elsewhere in the collection. These serve as \textit{semantic tags} for expanding $d_q$. 

In the \textbf{unsupervised} task setting described later, a document to be tagged is regarded as a query document $d_q$; its semantic tags are generated via PRF, and these terms will in turn be used for PRF--based expansion of unseen queries in downstream IR. Thus $d_q$ could represent an original complex query text or a document in the collection. 

In the following sections we describe the building blocks used in the setup for the baseline and proposed models for sequence-to-set semantic tagging as described in the task definition.

\subsection{Training and Inference Setup}
The overall architecture for sequence-to-set semantic tagging consists of two phases, as depicted in the block diagrams in Figures \ref{seq2set_fig}(a) and \ref{seq2set_fig}(b): the first, for training of input representations of documents; and the second for inference to achieve \textit{term transfer} for semantic tagging. As shown in Figure \ref{seq2set_fig}(a), the proposed model architecture would first learn the appropriate feature representations of documents in a first pass of training, by taking in the tokens
%\footnote{In our work, the tokens are different word-embedding schemes trained on our corpus, listed in Appendix \ref{wordembeddings}} 
of an input document sequentially, using a document's pre-determined $top-k$ TFIDF-scored terms as the \textbf{\textit{pseudo-}}class labels for an input instance, i.e. prediction targets for a $sigmoid$ layer for multi-label classification. The training objective is to maximize probability for these $k$ terms, i.e. $y_p = (t_1,t_2,..t_k) \in V$, i.e.
\begin{equation}
    %\arg max_{(y_p = (t_1,t_2,..t_k) \in V,\theta)} {P(y_p/v)}
    \arg \max_{\theta} P(y_p = (t_1,t_2,..t_k) \in V|v; \theta)
\end{equation}
given the document's encoding $v$. For computational efficiency, we take $V$ to be the list of top-10K TFIDF-scored terms from our corpus, thus $|V| = 10,000$. $k$ is taken as 3, so each document is initially labeled with 3 terms. The sequential model is then trained with the $k$-hot 10K-dimensional label vector as targets for the sigmoid classification layer, employing a couple of alternative training objectives. The first, typical for multi-label classification, minimizes a categorical cross-entropy loss, which for a single training instance with ground-truth label \textit{set}, $y_p$, is:
%given as: 
%\begin{equation}
%    L(\hat{y})= \sum_{p=1}^{k} log(\hat{y_p})
%\end{equation}
\begin{equation}
\label{xent}
    L_{CE}(\hat{y_p}) = \sum_{i=1}^{|V|} y_i \log(\hat{y_i})
\end{equation}
%\begin{samepage}
Since our goal is to obtain the most meaningful document representations most predictive of their assigned terms, and that can also be predictive of semantic tags not present in the document, we also consider a language model--based loss objective converting our decoder to a neural language model. Thus, we employ a training objective that maximizes the conditional log likelihood of the label terms $L_d$ of a document $d_q$, given the document's representation $v$, i.e. $P(L_d|d_q)$ (where $y_p = L_d \in V$). This amounts to minimizing the negative log likelihood 
%summation of the log probabilities 
of the label representations conditioned on the document encoding. %\end{samepage}
Thus,  
\begin{equation}
\label{lmeq1}
    P(L_d|d_q) = \prod\sb{l \in L_d} P(l|d_q) = - \sum\sb{l \in L_d}\log (P(l|d_q))
\end{equation}
Since $P(l|d_q) \propto \exp(v_l \cdot v)$, where $v_l$ and $v$ are the label and document encodings, %we have
it is equivalent to minimizing:
%Since $P(l|d_q) \propto \exp(vec(l) . vec(d_q))$, where $vec(l)= v_l$ and $vec(d_q) = v$, we have
%\begin{equation} 
%\label{lmeq2}
%    \prod\sb{l \in L_d} P(l|d_q) =  - \sum\sb{l \in L_d} \log(\exp(v_l . v))
%\end{equation}
%\begin{equation} %j li
%\label{lmeq2}
%    \prod\sb{l \in L_d} P(l|d_q) \propto - \sum\sb{l \in L_d} v_l \cdot v
%\end{equation}

%\begin{equation} 
%\label{lmeq2}
%    \prod\sb{l \in L_d} P(l|d_q) \propto - \sum\sb{l \in L_d} \log(\exp(v_l \cdot v))
%\end{equation}

\begin{equation} 
\label{lmeq2}
    %\prod\sb{l \in L_d} P(l|d_q) = 
    L_{LM}(\hat{y_p}) = - \sum\sb{l \in L_d} \log(\exp(v_l \cdot v))
\end{equation}

Equation (\ref{lmeq2}) represents our language model--style loss objective. We run experiments training with both losses  (Equations (\ref{xent}) \& (\ref{lmeq2})) as well as a variant that is a summation of both, with a hyper-parameter $\alpha$ used to tune the language model component of the total loss objective. 
%that has an advantage of propagating dense gradients as opposed to the sparse gradient updates

\subsection{Sequence-based Document Encoders} \label{encmodels}

We describe below the different neural models that we use for the sequence encoder, as part of our encoder-decoder architecture for deriving semantic tags for documents.

\subsubsection{doc2vec encoder} 

$doc2vec$ is the unsupervised algorithm due to \cite{le2014distributed}, that learns fixed-length representations of variable length documents, representing each document by a dense vector trained to predict surrounding words in contexts sampled from each document. We derive these doc2vec encodings by pre-training on our corpus. We then use them directly as features for inferring semantic tags per Figure \ref{seq2set_fig}(b) without training them within our framework against the loss objectives. We expect this to be a strong document encoding baseline in capturing the semantics of documents. TFIDF Terms is our other baseline where we don't train within the framework but rather use the top-$k$ neighbor documents' TFIDF pseudo-labels as the semantic tags for the query document.

\subsubsection{Deep Averaging Network encoder}

The Deep Averaging Network (DAN) for text classification due to \cite{iyyer2015deep} Figure \ref{seq2set_fig} (c), is formulated as a neural bag of words encoder model for mapping an input sequence of tokens $X$ to \textit{one} of $k$ labels. $v$ is the output of a composition function $g$, in this case $averaging$, applied to the sequence of word embeddings $v_w$ for $w \in X$. For our multi-label classification problem, $v$ is fed to a \textit{sigmoid} layer to obtain scores for each independent classification. We expect this to be another strong document encoder given results in the literature and it proves in practice to be.

\subsubsection{LSTM and BiLSTM encoders}\label{lstmbilstm}

% Long Short-Term Memory or LSTM recurrent neural networks are capable of learning and remembering over long sequences of inputs.
LSTMs \cite{hochreiter1997long}, by design, encompass memory cells that can store information for a long period of time and are therefore capable of learning and remembering over long and variable sequences of inputs. In addition to three types of gates, i.e. \textit{input}, \textit{forget}, and \textit{output} gates, that control the flow of information into and out of these cells, LSTMs have a hidden state vector $h_t^l$,  and a memory vector $c_t^l$. At each time step, corresponding to a token of the input document, the LSTM can choose to read from, write to, or reset the cell using explicit gating mechanisms. Thus the LSTM is able to learn a language model for the entire document, encoded in the hidden state of the final timestep, which we use as the document encoding to give to the prediction layer. By the same token, owing to the bi-directional processing of its input, a BiLSTM-based document representation is expected to be even more robust at capturing document semantics than the LSTM, with respect to its prediction targets. Here, the document representation used for final classification is the concatenated hidden state outputs from the final step, $[\overrightarrow{h}_t^l;\overleftarrow{h}_t^l]$, depicted by the dotted box in Fig. \ref{seq2set_fig}(e). 

\subsubsection{BiLSTM with Attention encoder}
In addition, we also propose a BiLSTM with attention-based document encoder, where the output representation is the \textit{weighted combination} of the concatenated hidden states at each time step. Thus we  learn an  \textit{attention-weighted} representation at the final output as follows.  Let $X \in \R^{(d \times L)}$ be a matrix consisting of output vectors $[h_1, \ldots, h_L]$ that the Bi-LSTM produces when reading $L$ tokens of the input document. Each word representation $h_i$ is obtained by concatenating the forward and backward hidden states, i.e. $h_i=[\overrightarrow{h_i}; \overleftarrow{h_i}]$. $d$ is the size of embeddings and hidden layers. The attention mechanism produces a vector $\alpha$ of attention weights and a weighted representation $r$ of the input, via:
\begin{equation}
    M = \tanh(W X), \quad M \in \R^{(d \times L)}
\end{equation}
\begin{equation}
	\alpha = \text{softmax}(w^T M), \quad \alpha \in \R^L
\end{equation}
\begin{equation}
	r = X {\alpha}^T, \quad r \in \R^d
\end{equation}

Here, the intermediate attention representation $m_i$ (i.e. the $i^{th}$ column vector in $M$) of the $i^{th}$ word in the input document is obtained by applying a non-linearity on the matrix of output vectors $X$, and the attention weight for the $i^{th}$ word in the input is the result of a weighted combination (parameterized by $w$) of values in $m_i$. Thus $r \in \R^d$ is the $attention-$weighted representation of the word and phrase tokens in an input document used in optimizing the training objective in downstream multi-label classification, as shown by the final attended representation $r$ in Figure \ref{seq2set_fig}(e).

\subsubsection{GRU and BiGRU encoders} \label{grubigru}
%[[Information about GRU and BIGRU goes here]]
A Gated Recurrent Unit (GRU) is a type of recurrent unit in recurrent neural networks (RNNs) that aims at tracking long-term dependencies while keeping the gradients in a reasonable range. In contrast to the LSTM, a GRU has only 2 gates: a reset gate and an update gate. First proposed by \cite{chung2014empirical}, \cite{chung2015gated} to make each recurrent unit to adaptively capture dependencies of different time scales, the GRU, however, does not have any mechanism to control the degree to which its state is exposed, exposing the whole state each time. In the LSTM unit, the amount of the memory content that is seen, or used by other units in the network is controlled by the output gate, while the GRU exposes its full content without any control. Since the GRU has simpler structure, models using GRUs generally converge faster than  LSTMs, hence they are faster to train and may give better performance in some cases for sequence modeling tasks. The BiGRU has the same structure as GRU except constructed for bi-directional processing of the input, depicted by the dotted box in Fig. \ref{seq2set_fig}(e).

\subsubsection{Transformer self-attentional encoder} \label{transformer}

Recently, the Transformer encoder-decoder architecture due to \cite{vaswani2017attention}, based on a \textit{self-attention} mechanism in the encoder and decoder, has achieved the state-of-the-art in machine translation tasks at a fraction of the computation cost. Based entirely on attention, and replacing the recurrent layers commonly used in encoder-decoder architectures with \textit{multi-headed} self-attention, it has outperformed most previously reported ensembles on the task. Thus we hypothesize that this self-attention-based model could learn the most efficient semantic representations of documents for our unsupervised task. %Since our models use $tensorflow$ \cite{abadi2016tensorflow}, a natural choice was document representation learning using the Transformer model's available $tensor2tensor$ API. 
We hoped to leverage apart from the computational advantages of this model, the capability of capturing semantics over varying lengths of context in the input document, afforded by multi-headed self-attention, Figure \ref{seq2set_fig}(f).  Self-attention is realized in this architecture, by training 3 matrices, made up of vectors, corresponding to a Query vector, a Key Vector and a Value vector for each token in the input sequence. The output of each self-attention layer is a summation of  weighted Value vectors that passes on to a feed-forward neural network. Position-based encoding to replace recurrences help to lend more parallelism to computations and make things faster. \textit{Multi-headed} self-attention further lends the model the ability to focus on different positions in the input, with multiple sets of Query/Key/Value weight matrices, which we hypothesize should result in the most effective document representation, among all the models, for our downstream task.

%\subsubsection{CNN encoder}

%Inspired by the success of \cite{kim2014convolutional} in employing CNN architectures successfully for achieving gains in NLP tasks we also employ a CNN-based encoder in the seq2set framework. \cite{kim2014convolutional} train a simple CNN with a layer of convolution on top of pre-trained word vectors, as a sequence of length $n$  embeddings concatenated to form a matrix input. Filters of different sizes, representing various context windows over neighboring words, are then applied to this input, over each possible window of words in the sequence to obtain feature maps. This is followed by a max-over-time pooling operation to take maximum value of the feature map as the feature corresponding to a particular filter. The model then combines these features to form a penultimate layer which is passed to a fully connected softmax layer whose output is the probability distribution over labels. In case of seq2set these features are passed to sigmoid layer for final multi-label prediction used cross entropy loss or a combination of cross-entropy and LM losses. We use filters of sizes 2, 3, 4 and 5. Like our other encoders, we fine-tune the document representations learnt.

\section{Task Settings}
%\section{Unsupervised Task Setting -- Semantic Tagging for Query Expansion}
\subsection{Semantic Tagging for Query Expansion}\label{unsupexp}

We now describe the setup and results for experiments run on our unsupervised task setting of semantic tagging of documents for PRF--based query expansion

\subsubsection{Dataset -- TREC CDS 2016}

The 2016 TREC CDS challenge dataset, 
%designed with the purpose of investigating techniques for linking medical records to information relevant for patient care, 
makes available actual electronic health records (EHR) of patients (de-identified), in the form of case reports, typically describing a challenging medical case. Such a case report represents a \textbf{query} in our system, having a complex information need. There are 30 queries in this dataset, corresponding to such case reports, at 3 levels of granularity \textbf{Note}, \textbf{Description} and \textbf{Summary} text as described in \cite{overviewcds2016}. The target document collection is the Open Access Subset of PubMed Central (PMC), containing 1.25 million articles consisting of \textbf{$title$}, \textbf{$keywords$}, \textbf{$abstract$} and \textbf{$body$} sections. We use a subset of 100K of these articles for which human relevance judgments are made available by TREC, for training. Final evaluation however is done on an ElasticSearch index built on top of the entire collection of 1.25 million PMC articles.

\subsubsection{Experiments -- Unsupervised Task Setting}

We ran several sets of experiments with various document encoders, employing word embedding schemes like skip-gram \cite{mikolov2013efficient} and Probabilistic Fasttext \cite{athiwaratkun2018probabilistic}. The experimental setup used is the same as the Phrase2VecGLM \cite{das2018phrase2vecglm}, the only other known system for this dataset, that performs ``unsupervised semantic tagging of documents by PRF'', for downstream query expansion. Thus we take this system as the current state-of-the-art system baseline, while our \textit{non-attention-based} document encoding models constitute our standard baselines. Our document-TFIDF representations--based query expansion forms yet another baseline. Summary text UMLS \cite{bodenreider2004unified} terms for use in our augmented models is available to us via the UMLS Java Metamap API similar to \cite{chenevaluation}. The first was a set of experiments with our different models using the \texttt{Summary Text} as the base query. Following this we ran experiments with our models using the \texttt{Summary Text + Sum. UMLS terms} as the ``augmented'' query. We use the Adam optimizer \cite{kingma2014adam} for training our models. After several rounds of hyper-paramater tuning, $batch\_size$ was set to $128$, $dropout$ to $0.3$, the prediction layer was fixed to $sigmoid$, the loss function switched between cross-entropy and summation of cross entropy and LM losses, and models trained with early stopping. %our models run between typically 100 to 200 epochs. 

Results from various \textit{Seq2Set} encoder models on \textbf{base} (\texttt{Summary Text})  and \textbf{augmented} (\texttt{Summary Text + Summary-based UMLS terms}) query, are outlined in Table \ref{tab1}. Evaluating on base query, a \textit{Seq2Set}-Transformer model beats other \textit{Seq2Set} encoders, and TFIDF, MeSH QE terms and Expert QE terms baselines; and a \textit{Seq2Set}-GRU model outperforms the Phrase2VecGLM baseline by ensemble with P@10 \textbf{0.3222}. On the augmented query, the \textit{Seq2Set}-BiGRU and \textit{Seq2Set}-Transformer models outperform other \textit{Seq2Set} encoders, and the Phrase2VecGLM unsupervised QE system baseline with P@10 of \textbf{0.4333}. Best performing \textit{supervised QE} systems for this dataset, tuned on all 30 queries, range between \textit{0.35}--\textit{0.4033} P@10
%, and \textit{0.25}--\textit{0.28} NDCG per
\cite{overviewcds2016}, better than unsupervised QE systems on base query, but surpassed by the best \textit{Seq2Set}-based models on augmented query even without ensemble. Semantic tags from a best-performing model, do appear to pick terms relating to certain conditions, e.g.: \textit{\textless \textbf{query\_doc original pseudo-label terms}:[\textquotesingle obesity\textquotesingle, \textquotesingle diabetes\textquotesingle, \textquotesingle pulmonary-hypertension\textquotesingle, \textquotesingle children\textquotesingle], \textbf{semantic tags}: [\textquotesingle dyslipidaemia\textquotesingle, \textquotesingle hyperglycemia\textquotesingle, \textquotesingle bmi\textquotesingle, \textquotesingle subjects\textquotesingle ] \textgreater.}

\begin{table}[h!]
\label{tab2}
\small
\centering
\begin{tabular}{|l|l|}
    \hline 
    %& & \\
    \bf %\textit{Unsupervised} QE Systems on Base Query %& \bf inf.  
    & \bf   \\ 
    \bf Unsupervised QE Systems -- Summary Text Query%(Model on Summary Text) %& \bf NDCG 
    & \bf P@10 \\ 
    %& & \\
    \hline
    BM25+\textit{Seq2Set}-doc2vec (\bf baseline)  %& 0.0134 
    &  0.0794 \\
    %\hline
    BM25+\textit{Seq2Set}-TFIDF Terms (\bf baseline)%& 0.0427 
    &  0.2000 \\
    %\hline
    BM25+\textbf{MeSH} QE Terms (\bf baseline)  %& 0.0970 
    & 0.2294 \\
    %\hline
    %BM25+None (\bf baseline)  %& 0.1057 
    %& 0.2489 \\
    %\hline
    BM25+\textbf{Human Expert} QE Terms 
    %& & \\  
    (\bf baseline) %& 0.1029 
    & 0.2511 \\
    %\hline
    %BM25+\textit{Seq2Set}-Transformer (Xent.+ LM) (\bf  model)   %& 0.0259 
    %& \bf 0.2667* \\  %%% removed this one
    %\hline
    BM25+unigramGLM+Phrase2VecGLM %& 
    & \\    
    \textit{ensemble} (\textbf{system baseline}) \cite{das2018phrase2vecglm}
    %(Das et al.) %& 0.1057 
    & \bf 0.2756 \\
    %\hline
    BM25+unigramGLM+Phrase2VecGLM+ %& 
    & \\ 
    \textit{Seq2Set}-GRU (LM only loss) \textit{ensemble} (\bf  model) %& 0.0957 
    & \bf 0.3222* \\ 
    \hline
    \bf Supervised QE Systems -- Summary Text Query%\textit{Supervised} QE Systems on Base Query %& \bf  
    & \bf  \\
    \hline
    BM25+ETH Zurich-\textit{ETHSummRR}  %& 0.2179 
    & 0.3067     \\
    BM25+Fudan Univ.DMIIP-\textit{AutoSummary1}  %& 0.2815 
    & \bf 0.4033     \\
    \hline    
    %\multicolumn{3}{|l|}{}\\
    %\bf \textit{Unsupervised} QE Systems on Augmented Query %&\bf 
    \bf Unsupervised QE Systems -- Summary Text  & \bf \\
    \bf + Summary--based UMLS Terms Query & \\
    %\bf (Model on Summary+Sum. UMLS terms) %& \bf & \bf \\
    \hline
    BM25+\textit{Seq2Set}-doc2vec (\bf baseline)  %& 0.0441 
    & 0.1345 \\ 
    %\hline
    %\hline
    BM25+\textit{Seq2Set}-TFIDF Terms (\bf baseline)   %& 0.1119 
    &  0.3000 \\ 
    %\hline
    %BM25+\textit{Seq2Set}-BiLSTM w/ Attn. (Xent.)
    %& & \\ 
    %(\bf model) %& 0.0608 
    %& 0.3000 \\ 
    %\hline
    BM25+unigramGLM+Phrase2VecGLM %& 
    & \\
     \textit{ensemble} (\textbf{system baseline}) \cite{das2018phrase2vecglm}
     %(Das et al.) %& 0.1206 
     & \bf 0.3091 \\
    %\hline
    BM25+\textit{Seq2Set}-BiGRU (LM only loss) (\bf model)  %&  0.1340 
    & \bf 0.3333* \\
    %\hline
    %BM25+unigramGLM+Phrase2VecGLM+ %& 
    %& \\
    %\textit{Seq2Set}-Transformer (Xent.+ LM) \textit{ensemble} 
    %& & \\
    %(\bf model)  %&  0.1194 
    %& \bf 0.4167  \\
    %\hline
    %BM25+unigramGLM+Phrase2VecGLM+ %& 
    %& \\
    %\textit{Seq2Set}-BiGRU (LM only loss) \textit{ensemble}
    %& & \\ %%% removed this one
    %(\bf model)  %&  \bf 0.1441 
    %& \bf 0.4167 \\
    BM25+\textit{Seq2Set}-Transformer (Xent.+ LM) (\bf model)   %& 0.0474 
    &  \bf 0.4333* \\
    \hline
\end{tabular}
\caption{\label{tab1} Results on IR for best \textit{Seq2set} models, in an \textbf{\textit{unsupervised PRF}}--based QE setting. Boldface indicates statistical significance @p$<<$0.01 over previous.}
%result. inf. is for inferred measure \cite{yilmaz2008simple} }
%\footnote{The inf. prefix on the metrics indicate \textit{inferred} measures used for this task \cite{voorhees2014effect}.
\end{table}

\subsection{Automated Text Categorization -- Supervised} \label{supexp}

The Seq2set framework's unsupervised semantic tagging setup is primarily applicable in those settings where no pre-existing document labels are available. In such a scenario, of unsupervised semantic tagging of a large document collection, the Seq2set framework therefore consists of separate training and inference steps to infer tags from other documents after encodings have been learnt.  We therefore conduct a series of extensive evaluations in the manner described in the previous section, using a downstream QE task in order to validate our method. However, when a tagged document collection is available where the set of document labels are already known, we can learn to predict tags from this set of known labels on a new set of similar documents. Thus, in order to generalize our Seq2set approach to such other tasks and setups, we therefore aim to validate the performance of our framework on such a labeled dataset of tagged documents, which is equivalent to adapting the Seq2set framework for a supervised setup. In this setup we therefore only need to use the training module of the Seq2set framework shown in Figure \ref{seq2set_fig}(a), and measure tag prediction performance on a held out set of documents. For this evaluation, we therefore choose to work with the popular Delicious (del.icio.us) folksonomy dataset, same as that used by \cite{soleimani2016semi} in order to do an appropriate comparison with their framework that is also evaluated on a similar document multi-label prediction task.

\subsubsection{Dataset -- del.icio.us}

The Delicious dataset contains tagged web pages retrieved from the social bookmarking site, del.icio.us. %As in \cite{ramage2011partially}, 
There are 20 common tags used as class labels: $reference$, $design$, $programming$, $internet$, $computer$, $web$, $java$, $writing$, $English$, $grammar$, $style$, $language$, $books$, $education$, $philosophy$, $politics$, $religion$, $science$, $history$ and $culture$. The training set consists of 8250 documents and the test set consists of 4000 documents. 

\subsubsection{Experiments -- Supervised Task Setting}

We then run Seq2set-based training for our 8 different encoder models on the training set for the 20 labels, and perform evaluation on the test set  measuring sentence-level ROC AUC on the labeled documents in the test set.

\begin{figure}[t!]
    \centering
    \includegraphics[width=0.4\textwidth]{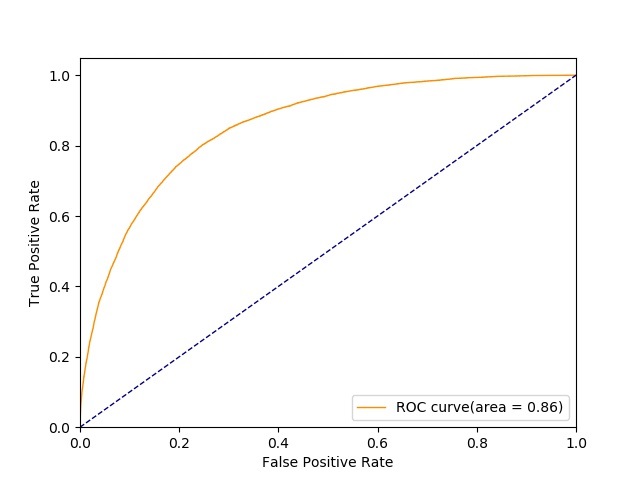}
    %\caption{Seq2Set on \textbf{del.icio.us} -- BiGRU--based encoder}
    \caption{Seq2Set--\textbf{supervised} on \textbf{del.icio.us}, best model--BiGRU--based encoder}
    %\caption{Seq2Set -- \textbf{supervised} text categorization task setting ROC AUC on \textbf{del.icio.us} dataset for best performing models}
 \label{deliciousaucfig}
\end{figure}

Figure \ref{deliciousaucfig} shows the ROC AUC for the best performing model from the Seq2set framework on the del.icio.us dataset, %. Figures \ref{fig:1}--\ref{fig:3} show an AUC of 0.85 obtained with our best performing, GRU, BiLSTM with Attention and Transformer--based models respectively while our very best on multi-label prediction with the del.icio.us dataset 
which was due to a BiGRU--based encoder model trained with a sigmoid--based prediction layer on cross entropy loss with a batch size of 64 and dropout set to 0.3. This best model got an ROC AUC of \textbf{0.86 }, very significantly surpassing MLTM (AUC 0.81) for this task and dataset.

\begin{figure}[t!]
    \centering
    \includegraphics[width=0.4\textwidth]{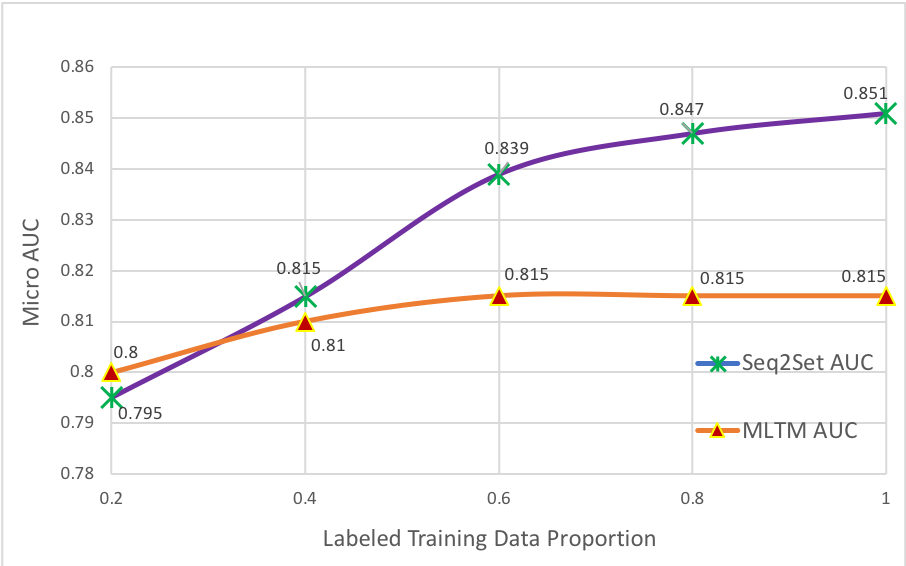}
    \caption{A comparison of document labeling performance of Seq2set versus MLTM }
    \label{auc_del_mltm_comp}
\end{figure}

Figure \ref{auc_del_mltm_comp} also shows a comparison of the ROC AUC scores obtained with training Seq2set and MLTM based models for this task with various labeled data proportions. Here again we see that Seq2set has clear advantage over the current MLTM state-of-the-art, statistically significantly surpassing it ($p << 0.01$) when trained with greater than ~25\% of the labeled dataset.
%\begin{figure}
%  \begin{subfigure}[b]{0.33\textwidth}
%    \includegraphics[width=\textwidth]{./1Seq2setbestGRUdelicious.png}
%    \caption{Seq2Set on \textbf{del.icio.us} -- GRU--based encoder}
%    \label{fig:2.1}
%  \end{subfigure}
%  %
%  \begin{subfigure}[t]{0.33\textwidth}
%    \includegraphics[width=\textwidth]{./3Seq2setbestBiLSTMATTdelicious.png}
%    \caption{Seq2Set on \textbf{del.icio.us} -- BiLSTM with Attention--based encoder}
%    \label{fig:2.2}
%  \end{subfigure}
%
%  \begin{subfigure}[b]{0.33\textwidth}
%    \includegraphics[width=\textwidth]{./4Seq2setbestTransformerdelicious.png}
%    \caption{Seq2Set on \textbf{del.icio.us} -- Transformer--based encoder}
%    \label{fig:2.3}
%  \end{subfigure}
% 
%  \begin{subfigure}[b]{0.33\textwidth}
%    \includegraphics[width=\textwidth]{./0Seq2setbestBiGRUdelicious.png}
%    \caption{Seq2Set on \textbf{del.icio.us} -- BiGRU--based encoder}
%    \label{fig:2.4}
%  \end{subfigure}
%  \caption{Seq2Set -- \textbf{supervised} text categorization task setting ROC AUC on \textbf{del.icio.us} dataset for best performing models}
% \label{deliciousaucfig}
%\end{figure}

%\newpage

\subsection{Automated Text Categorization -- Semi-Supervised} \label{semisupexp}

We then seek to further validate how well the Seq2set framework can leverage large scale pre-training on unlabeled data given only a small amount of labeled data for training, to be able to improve prediction performance on a held out set of these known labels. This amounts to a semi-supervised setup--based evaluation of the Seq2set framework. In this setup, we perform the training and evaluation of Seq2set similar to the supervised setup, except we have an added step of pre-training the multi-label prediction on large amounts of unlabeled document data in exactly the same way as the unsupervised setup. 

\subsubsection{Dataset -- Ohsumed}

We employ the Ohsumed dataset available from the TREC  Information Filtering tracks of years 87-91 and the version of the labeled \textbf{Ohsumed} dataset used by \cite{soleimani2016semi} for evaluation, to have an appropriate comparison with their MLTM system also evaluated for this dataset. The version of the Ohsumed dataset due to \cite{soleimani2016semi} consists of 11122 training and 5388 test documents, each assigned to one or multiple labels of  23 MeSH diseases categories. Almost half of the documents have more than one label.  %description about Ohsumed goes here% 

\begin{figure}[t!]
    \centering
    \includegraphics[width=0.4\textwidth]{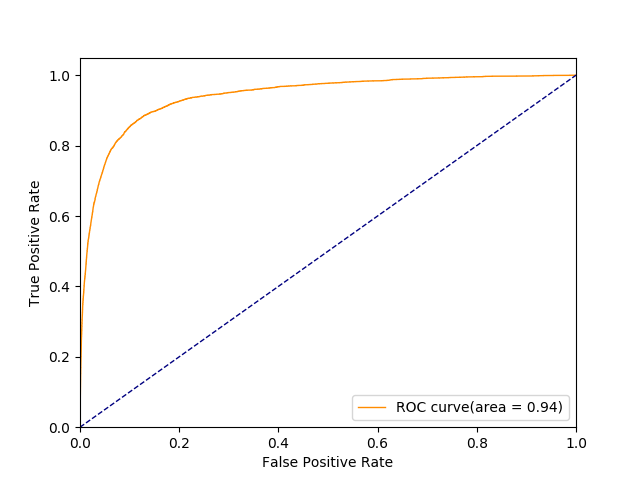}
    \caption{Seq2Set--\textbf{semi-supervised} on \textbf{Ohsumed}, best model--Transformer encoder w/ Cross Entropy--based Softmax prediction; 4 layers, 10 attention heads, dropout=0}
    %  \caption{Seq2Set -- supervised and \textbf{semi-supervised} text categorization task setting ROC AUC on \textbf{Ohsumed} dataset for best performing models}
  \label{ohsumedaucfig}
\end{figure}

%\begin{figure}
%  \begin{subfigure}[t]{0.33\textwidth}
%    \centering
%    \includegraphics[width=\textwidth]{./1_BiGRU_softmax_uniform.png}
%    \caption{Seq2Set; supervised on \textbf{Ohsumed} -- BiGRU encoder with Cross Entropy--based Softmax prediction}
%    \label{fig:1}
%  \end{subfigure}
%  \begin{subfigure}[t]{0.33\textwidth}
%    \includegraphics[width=\textwidth]{./2_CNN_softmax_uniform.png}
%    \caption{Seq2Set; supervised on \textbf{Ohsumed} -- CNN encoder with Cross Entropy--based Softmax prediction}
%    \label{fig:2}
%  \end{subfigure}
%  \begin{subfigure}[b]{0.33\textwidth}
%    \includegraphics[width=\textwidth]{./3_Transformer_nh10_sigmoid.png}
%    \caption{Seq2Set; supervised on \textbf{Ohsumed} -- Transformer encoder with Cross Entropy--based Sigmoid prediction; 4 layers, 10 attention heads, dropout=0.2 }
%    \label{fig:3}
%  \end{subfigure}
%  \begin{subfigure}[b]{0.33\textwidth}
%    \includegraphics[width=\textwidth]{./4_Transformer_nh10_softmax_uniform.png}
%    \caption{Seq2Set; \textbf{semi-supervised} on \textbf{Ohsumed} -- Transformer encoder with Cross Entropy--based Softmax prediction; 4 layers, 10 attention heads, dropout=0}
%    \label{fig:4}
%  \end{subfigure}
%  \caption{Seq2Set -- supervised and \textbf{semi-supervised} text categorization task setting ROC AUC on \textbf{Ohsumed} dataset for best performing models}
%  \label{ohsumedaucfig}
%\end{figure}

\subsubsection{Experiments -- Semi-Supervised Task Setting}

We first train and test our framework on the labeled subset of the \textbf{Ohsumed} data from \cite{soleimani2016semi} similar to the supervised setup described in the previous section. This evaluation gives a statistically significant ROC AUC of 0.93 over the 0.90 AUC for the MLTM system of \cite{soleimani2016semi} for a  Transformer--based Seq2set model performing best. Next we experiment with the  semi-supervised setting where we first train the Seq2set framework models on a large number of documents that do not have pre-existing labels. This pre-training is performed in exactly a similar fashion as the unsupervised setup. Thus we first preprocess the Ohsumed data from years 87-90 to obtain a top-1000 TFIDF score--based vocabulary of tags, pseudo-labeling all the documents in the training set with these. Our training and evaluation for the semi-supervised setup consists of 3 phases: \textbf{Phase 1}: We employ our seq2set framework (using each one of our encoder models) for multi-label prediction on this pseudo-labeled data, having an output prediction layer of 1000 having a penultimate fully-connected layer of dimension 23, same as the number of labels in the Ohsumed dataset; \textbf{Phase 2}: After pre-training with pseudolabels we discard the final layer and continue to train labeled Ohsumed dataset from 91 by 5-fold cross-validation with early stopping. \textbf{Phase 3}: This is the final evaluation step of our semi-supervised trained Seq2set model on the labeled Ohsumed test dataset used by \cite{soleimani2016semi}. This constitutes simply inferring predicted tags using the trained model on the test data. As shown in Figure \ref{ohsumedaucfig}, our evaluation of the Seq2set framework for the Ohsumed dataset, comparing supervised and semi-supervised training setups, yields an ROC AUC of \textbf{0.94} for our best performing \textbf{semi-supervised}--trained model of Fig. \ref{ohsumedaucfig}, compared to the various supervised trained models for the same dataset with a best ROC AUC of ~0.93. The best performing semi-supervised model again involves a Transformer--based encoder using a softmax layer for prediction, with 4 layers, 10 attention heads, and no dropout. Thus, the best results on the semi-supervised training experiments (ROC AUC 0.94) \textbf{statistically significantly outperforms} ($p << 0.01$) the MLTM baseline (ROC AUC ~0.90) on the Ohsumed dataset. %, it only slightly surpasses the best supervised Seq2set models. %However, we believe that more experiments with the Seq2set framework could yield an even better performing semi-supervised--trained model, that very significantly outperforms the supervised task for the same dataset.

\section{Conclusion} 
We develop a novel \textit{sequence-to-set} end-to-end encoder-decoder--based neural framework for multi-label prediction by training document representations using no external supervision labels, for pseudo-relevance feedback--based  unsupervised semantic tagging of a large collection of documents. We find that in this unsupervised task setting of PRF-based semantic tagging for query expansion, a multi-term prediction training objective that jointly optimizes both prediction of the TFIDF--based document pseudo-labels and the log likelihood of the labels given the document encoding, surpasses previous methods such as Phrase2VecGLM \cite{das2018phrase2vecglm} that used neural generalized language models for the same. Our initial hypothesis that bi-directional or self-attentional models could learn the most efficient semantic representations of documents when coupled with a loss more effective than cross-entropy at reducing language model perplexity of document encodings, is corroborated in all experimental setups. %We demonstrate the effectiveness of our approach in each task setting, i.e. for the unsupervised setting in a downstream medical IR challenge task, achieving to the best of our knowledge, the state-of-the-art on \textit{unsupervised} QE via PRF-based semantic tagging surpassing Phrase2VecGLM \cite{das2018phrase2vecglm}; and for both, supervised and semi-supervised settings, where Seq2set statistically significantly outperforms the state-of-art MLTM baseline \cite{soleimani2016semi} on the same held out set of documents as MLTM, for multi-label prediction on a set of known labels, for automated text categorization. %Thus our novel framework is effective for both unsupervised semantic tagging of documents as well as automated document categorization with known labels.
We demonstrate the effectiveness of our novel framework in every task setting, viz. for \textbf{unsupervised} QE via PRF-based semantic tagging for a downstream medical IR challenge task; as well as for both, \textbf{supervised} and \textbf{semi-supervised} task settings, where Seq2set statistically significantly outperforms the state-of-art MLTM baseline \cite{soleimani2016semi} on the same held out set of documents as MLTM, for multi-label prediction on a set of known labels, for automated text categorization; achieving to the best of our knowledge, the current state-of-the-art for multi-label prediction on documents, with or without known labels. %We thus demonstrate the effectiveness of our framework for multi-label prediction on documents, with or without labels.

%\clearpage
\bibliographystyle{aaai}
\bibliography{aaai}

\end{document}